\documentclass[11pt]{article}

\usepackage{xcolor,fullpage}
\usepackage{ulem}
\usepackage{pslatex}
\usepackage{graphicx}
\usepackage{fancyhdr}
\usepackage{amsmath}
\usepackage{amssymb}
\usepackage{textcomp}
\usepackage{natbib}	
\usepackage{subcaption}	
\usepackage{hyperref}
\usepackage{setspace}
\usepackage{cancel}
\usepackage{mathtools}
\usepackage[font=footnotesize]{caption}
\usepackage{subcaption}
\usepackage{lscape}
\usepackage{rotating}
\usepackage{adjustbox}
\usepackage{multirow}
\usepackage{tabularx}
\usepackage{lipsum}
\usepackage{booktabs}
\usepackage{subfiles}
\usepackage{titling}
\usepackage{placeins}
\usepackage{authblk}
\usepackage{comment}

\newcolumntype{L}[1]{>{\raggedright\let\newline\\\arraybackslash\hspace{0pt}}m{#1}}
\newcolumntype{C}[1]{>{\centering\let\newline\\\arraybackslash\hspace{0pt}}m{#1}}
\newcolumntype{R}[1]{>{\raggedleft\let\newline\\\arraybackslash\hspace{0pt}}m{#1}}

\title{Deep Learning of Causal Structures in High Dimensions}

\author[1]{Kai Lagemann*}
\author[2]{Christian Lagemann}
\author[3]{Bernd Taschler}
\author[1,4]{Sach Mukherjee*}
\affil[1]{\small Statistics and Machine Learning, DZNE, Bonn, Germany}
\affil[2]{Institute of Aerodynamics, RWTH Aachen University,  Aachen, Germany}
\affil[3]{Wellcome Centre for Integrative Neuroimaging, University of  Oxford, Oxford, UK }
\affil[4]{MRC Biostatistics Unit, University of Cambridge, Cambridge, UK}
\affil[*]{corresponding emails: $\{$kai.lagemann, sach.mukherjee$\}$@dzne.de}

\begin{document}
\date{}
\maketitle

\vspace{-2cm}

\begin{abstract}
\normalsize
\noindent
Recent years have seen rapid progress at the intersection between causality and machine learning. 
Motivated by scientific applications involving high-dimensional data, in particular in biomedicine,
we propose a deep neural architecture for learning causal 
relationships between variables from a combination of empirical data and prior causal knowledge.
We combine convolutional and graph neural networks 
within a causal risk framework
to 
provide a flexible and scalable  approach.
Empirical results include linear and nonlinear simulations (where the underlying causal structures are known and can be directly compared against), as well as a real biological example where the models are applied to high-dimensional molecular data and their output compared against entirely unseen validation experiments. These results demonstrate the feasibility of using  deep learning approaches to learn causal networks in large-scale problems spanning thousands of variables.

\end{abstract}

\section{Introduction}

 Causality remains an important open area in 
machine learning, statistics and related fields
  \citep[see e.g.][]{Peters2017,Arjovsky2019} and 
the task of identifying 
 causal relationships between variables
is key in many scientific domains 
including in particular biomedicine \citep[see e.g.][]{glymour2016causal,Hill2016}.
  The rich body of work in learning causal structures  includes, among other methods, PC \citep{Spirtes2000}, LiNGAM \citep{Shimizu2006}, IDA \citep{Maathuis2009}, GIES \citep{Hauser2012}, RFCI \citep{Colombo2012}, ICP \citep{Peters2015b} and MRCL \citep{Hill2019}. 
 However,  learning causal structures from data 
remains  
 challenging, particularly under conditions -- such as high dimensionality, limited data sizes, 
 presence of hidden variables etc. -- seen in many real-world problems.

In this paper, we propose a deep architecture for  causal learning that is motivated in particular by questions involving high-dimensional biomedical data. 
The approach we put forward operates within a paradigm 
that views causal questions through the lens of expected loss or risk (see below). 
The learners proposed allow for the integration of partial knowledge concerning a subset of causal relationships and then seek to generalize beyond what is initially known to learn relationships between all observed variables. 
This corresponds to a common scientific use-case, in which some prior knowledge is  available at the outset -- from previous experiments or scientific background knowledge -- but where the aim is to go beyond what is known to learn a model spanning all available variables. 

Much of the literature in learning causal structures involves statistical formulations that allow explicit description of the relevant data-generating distributions (including both observational and interventional distributions) and are in that sense ``generative" \citep[see, e.g.,][and references therein]{Heinze2018}. Taking a different approach, a number of recent papers, including \citet{lopez2015,Mooij2016,Hill2019,noe2019}, have considered learning discrete indicators of causal relationships between variables (without necessarily learning full details of the underlying data-generating models) and this is related to  notions of causal expected loss or risk \citep{eigenmann2020}. Such indicators may encode for example, whether, for a pair of variables $A$ and $B$, $A$ has a causal influence on $B$, $B$ on $A$, or neither. 

The approach we propose, called ``Deep Discriminative Causal Learning'' (D\textsuperscript{2}CL), 
is in the latter vein.
We consider a version of the causal structure learning problem in which the desired output consists of binary indicators of causal relationships between observed variables \citep{Hill2019, eigenmann2020}, which can be represented as a directed graph with nodes corresponding to the variables.  Available multivariate data $X$ are transformed to provide inputs to a neural network whose outputs are estimates of the causal indicators. 
As detailed below, D\textsuperscript{2}CL has several  differences to classical causal structure learning (e.g.\ based on causal graphical models). First, 
the  objective is different: rather than giving access to all interventional distributions, D\textsuperscript{2}CL outputs indicators of causal links. 
Second, D\textsuperscript{2}CL is highly non-parametric, relying on the learners  to detect relevant regularities.
Third, D\textsuperscript{2}CL is demonstrably scalable to large numbers of variables (and is in fact unsuitable for  small problems spanning only a few variables, see Discussion).
The assumptions underlying the approach  are also different in nature from the kinds of assumptions usually made in causal structure learning and concern higher-level regularities in the data-generating processes, as discussed further below.

 The remainder of the paper is organized as follows. We first introduce the D\textsuperscript{2}CL
methodology. 
 We then present  empirical results, on both synthetic, gold-standard problems and on real molecular biological data. In the latter case, model results are systematically checked against entirely unseen interventional experiments. Finally, we discuss open questions and limitations.

\section{Methods} \label{Section: The problem setting}
We propose  an end-to-end neural approach to learn causal networks from a combination of empirical data $X$ and prior causal knowledge $\Pi$. 
In this Section, we describe the proposed methodology, starting with notation and a problem statement and going on 
to present the learning scheme and architecture.

\subsection{Notation}
Observed variables with index set $V = \{1, \ldots, p \}$ are denoted $X_1, \ldots, X_p$. 
The variables will be identified with vertices in a directed graph $G$ whose vertex and edge sets are denoted $V(G),E(G)$, respectively. We occasionally overload $G$ to refer also to the corresponding binary adjacency matrix, using $G_{ij}$ to refer to the entry $(i,j)$ of the  adjacency matrix, as will be clear from context.
Where needed to make the distinction clear we will use $G^*$ to denote a true (unknown) graph  and $\hat{G}$ an estimate thereof. 
We use linear indexing of variable pairs
to aid formulation as a machine learning problem. Specifically, an ordered pair 
$(i,j) \in V \times V$ has an associated linear index $k \in \mathcal{K} = \{ 1, \ldots, K \}$, where $K$ is the total number of variable pairs of interest.
Where useful we make the mapping explicit, denoting
the linear index corresponding to a pair $(i,j)$ as $k(i,j)$ and the variable pair corresponding to a linear index $k$ as $(i(k), j(k))$. The linear indices of pairs whose causal relationships are {\it unknown} and of interest are $\mathcal{U} \subset \mathcal{K}$ and those pairs known in advance via input knowledge $\Pi$ are $\mathcal{T}(\Pi) \subset \mathcal{K}$ (the notation emphasizes the fact that the set $\mathcal{T}$ is, in general, determined by the input knowledge $\Pi$). In all  experiments $\mathcal{T}(\Pi)$ and $\mathcal{U}$ are disjoint, i.e., no prior causal information is available on the pairs $\mathcal{U}$ of interest.

\subsection{Problem statement} 
We focus on the setting in which available inputs are:
\begin{itemize}
\item[] (I1) Empirical data: an $n \times p$ data matrix $X$ whose columns correspond to variables $X_1, \ldots, X_p$. 
\item[] (I2) Causal background knowledge $\Pi$ providing information on a subset $\mathcal{T}(\Pi) \subset \mathcal{K}$ of causal relationships. \end{itemize}

For (I2), we assume that 
the prior  knowledge $\Pi$ can be viewed as 
information concerning the  causal status of a subset of variable pairs. That is, for some variable pairs $(X_i,X_j)$ the correct binary indicator $G^*_{ij}$, representing the presence/absence of an edge in the target graphical object, is provided as an input. 
In terms of linear indexing, these can be viewed as available ``labels" of causal status for the pairs $\mathcal{T}(\Pi) \subset \mathcal{K}$. No specific assumption is made on the data $X$, but in line with our focus on generalizing to unseen causal relationships, it is assumed that it does {\it not} contain interventional data corresponding to the pairs in $\mathcal{U}$. Furthermore, in all experiments, not only are the sets $\mathcal{T}$ and $\mathcal{U}$ disjoint, but we enforce the stronger requirement that $u \in \mathcal{U} \implies \nexists j : k(i(u),j) \in \mathcal{T}$,  meaning all interventions on which  models are tested are entirely novel, i.e.\ unrepresented in the inputs to the learner.

Thus, the learning task can be formulated as follows: given the inputs (I1) and (I2), the goal is to estimate for each ordered pair of variables $(X_i,X_j)$ with unknown causal relationship, whether or not $X_i$ has a causal influence on $X_j$, or equivalently to learn the underlying graph $G^*$.




\subsection{Summary of learning scheme}
With the notation above, the goal is to learn a graph 
whose nodes correspond to the  variables $X_1, \ldots, X_p$ and edges represent causal relationships.
To this end, we train a parameterized network $F_\theta$, i.e.  a nonlinear function $F$ with a set of unknown, trainable parameters $\theta$. 
This is possible since we know for each pair $k \in \mathcal{T}$ the causal status $G^*_{i(k), j(k)}$ based on input information $\Pi$.
The architecture we use as $F_\theta$ is detailed below, but for now assume this has been specified. Then, given the data $X$ and prior input $\Pi$, we learn parameters $\hat{\theta}(X,\Pi)$ under a loss that is supervised by the (causal) inputs/labels $Y_k = G^*_{i(k),j(k)}$ for all pairs $k \in \mathcal{T}(\Pi)$.
In contrast to MRCL \citep{Hill2019}, which is semi-supervised and does not scale to high-dimensions, our approach is supervised and aimed at high-dimensional problems and unlike \citet{noe2019} we use a deep learning framework that learns causally-informed embeddings. We share with \citet{eigenmann2020} an emphasis on causal risk, but our focus is on learning, rather than risk estimation.


At this stage, the trained network $F_{\hat{\theta}(X,\Pi)}$
allows assignment of causal status to {\it any} pair since it gives an estimate of the entire  graph including those pairs whose causal status was unknown. Specifically, the output is given by:
\begin{equation}
    \hat{G}_{ij} (X,\Pi)= 
    \begin{cases}
        F_{\hat{\theta}(X,\Pi)}(i,j; \, X)  & \text{if } k(i,j) \notin \mathcal{T}(\Pi)\\
        Y_{k(i,j)}(\Pi) & \text{otherwise}
    \end{cases}
    \label{eq:graph_supervised}
\end{equation}

\noindent
where $(i,j)$ are ordered variable pairs.
Note that the overall estimate depends solely on the data $X$ and causal information $\Pi$. By default, no change is made for pairs $\mathcal{T}$ whose status was known at the outset.
\citet{eigenmann2020} studied causal notions of risk  based on loss functions of the form $L(\hat{G}, G^*)$ that compare a graph estimate $\hat{G}$  with ground-truth $G^*$. 
In our setting, we consider a classification-type loss on the variable pairs $k$, where  the  causal status of known pairs $\mathcal{T}(\Pi)$ provides the training ``labels".
We therefore use the corresponding binary cross-entropy loss,  augmented by additional terms that, for instance, prevent exploding weights.

In the D\textsuperscript{2}CL framework 
the notion of causal influence 
encoded by the edges 
is rooted in the application setting and input information $\Pi$, since causal semantics are inherited via the problem setting rather than specified by a generative model  (see \citet{Hill2019} for  related discussion). Indeed, in the experiments below we show examples in which
D\textsuperscript{2}CL is used to  learn either direct or indirect/ancestral causal relationships,
depending on the setting and inputs.
We direct the interested reader to  
Appendix A for further discussion of assumptions.

\subsection{Architecture details}

{\bf CNN Tower:}
To capture distributional information from  empirical data $X$, a preprocessing step is required. In principle, this could be done via a variety of multi-dimensional transformations of $X$. We consider the simplest possible case, namely for a pair $(i,j)$ to consider only the corresponding columns $i$ and $j$ in the data matrix $X$. Specifically, we use the $n \times 2$ submatrix $X_{({\cdot, [i j])}}$, to form a bivariate kernel density estimate $f_{ij} = \mathrm{KDE}(X_{({\cdot, [i j])}})$. Note that this is in general asymmetric in the sense that $f_{ij} \neq f_{ji}$, which is important since we want to learn ordered/directed relationships. Evaluations of the KDE at equally spaced grid points on the plane (i.e.\ numerical values from the induced density function) are treated as the input to the CNN. The KDE itself is a standard bivariate approach using automated bandwidth selection following \cite{Silverman86}, \cite{turlach1993bandwidth}. This provides an ``image" of the data and allows us to leverage standard tools from computer vision. Furthermore, we concatenate channelwise the numerical KDE values on the regularly spaced grid with a positional encoding of the grid points.   

The specific network architecture of our CNN tower is inspired by a ResNet-54 architecture \citep{he2015deep}. From a high level perspective, it consists of a stem, five stages with $[3, 4, 6, 3, 3]$ ResNet blocks and multiple fully connected layers that transform the high-level feature maps into a latent space that is merged with the output of the GNN tower. The first ResNet block at each stage downsamples the spatial dimensions of the output of the previous stage by a factor of two. To enhance the computational efficiency of the bottleneck layers in each ResBlock, channel down- and up-sampling exploiting $1 \times 1$ convolutions is performed before and after each feature extraction CNN layer \citep{szegedy2014going}. We replaced ReLU activations by the parametric counterpart PReLU \citep{he2015delving}. 
Following \citet{xie2017aggregated},  we chose a full pre-activation of the convolutional layers,  normalization-activation-convolution.

\begin{figure}[t]
\centering
\includegraphics[width=0.85\textwidth]{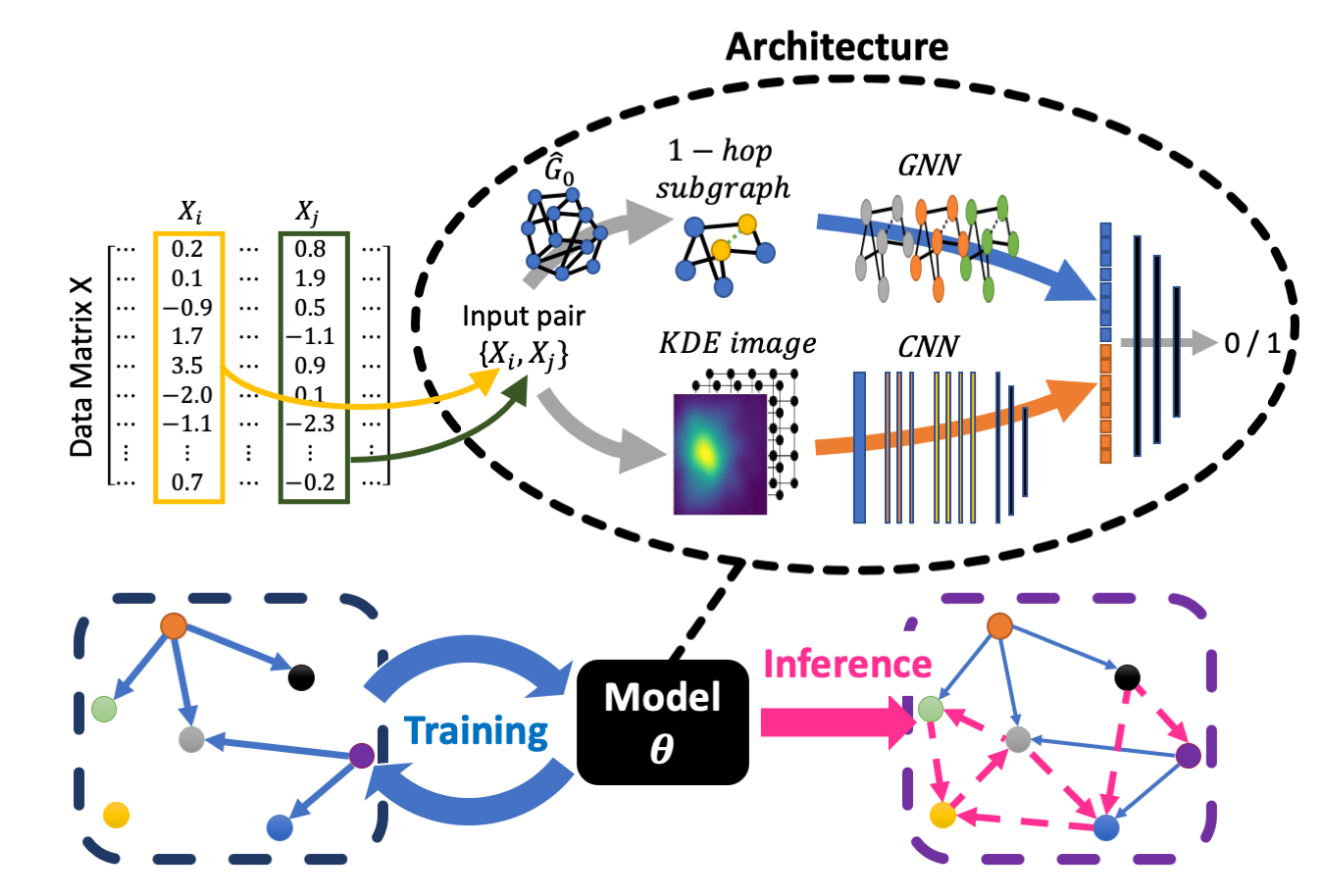}
\caption{Overview of the D\textsuperscript{2}CL 
architecture, training and inference. 
D\textsuperscript{2}CL combines empirical data with prior causal knowledge to learn causal relationships between variables. 
This is done using a neural architecture with two components: a CNN tower aimed at learning distributional features and a GNN tower that detects structural regularities. 
The CNN and GNN embeddings are then merged through multiple layers to estimate the probability of a directed causal relationship. 
During inference the network generalizes beyond the initial inputs to provide a global estimate spanning all variables of interest.}
\label{Figure: Architecture}
\end{figure}

\medskip
{\bf GNN tower.} The GNN tower leverages the SEAL architecture of \citet{zhang2018} and the resulting graph convolutional neural network (GCNN) for link prediction. The underlying notion is that a heuristic function predicts scores for the existence of a link. However, instead of employing predefined heuristics (such as the Katz coefficient or PageRank), an adaptive  function is learned in an end-to-end fashion, which is formulated as a graph classification problem on enclosing subgraphs.
Here, for a node pair of interest $(i, j)$, the GNN tower is intended to learn causally relevant node features and state embeddings based on a local 1-hop enclosing subgraph  extracted from an initial input graph $\hat{G}_0$. This is done as follows. For node pair $(i,j)$, we first extract a set $\mathcal{N}$ of neighbouring nodes  comprising all nodes connected to either $i$ or $j$ in $\hat{G}_0$. Then, the edge structure within the subgraph $G_{ij}$ is reconstructed by pulling out all edges from $\hat{G}_0$ for which the parent and child node are in $\mathcal{N}$. The order of the nodes is shuffled for each subgraph. The node features in every input subgraph consist of structural node labels that are assigned by a \textit{Double-Radius Node Labeling} (DRNL) heuristic \citep{zhang2018} and the individual data features. In a first step, the distances between node $i$ and all other nodes of the local subgraph except node $j$ are computed. The same is repeated for node $j$. A hashing function then transforms the two distance labels into a DRNL label that assigns the same label to nodes that are on the same ``orbit'' around the center nodes $i$ and $j$. During the training process the DRNL label is transformed into a one-hot encoded vector and passed to the first graph convolutional layer. In contrast to traditional CNNs, GCNNs do not benefit strongly from very deep architecture design \citep{chen2019,Li2018}. Therefore, our GNN tower consists only of four sequentially stacked graph convolutional layers. The activation function is the hyperbolic tangent.  Since the number of nodes in the enclosing subgraph for each pair of variables $(i, j)$ is different, a SortPooling layer \citep{Zhang2018AnED} is applied to select the top $k$ nodes according to their structural role within the graph. Afterwards, 1-dimensional convolutions extract features from the selected state embeddings.

{\bf Embedding Fusion.} Each tower outputs an embedding; these are concatenated and further processed by multiple fully connected layers. Finally, the last layers output the log-likelihood of a directed edge from node $i$  to node $j$. 

{\bf Implementation summary.}
All network architectures were implemented in the open source framework PyTorch \citep{paszke2019Pytorch}. The GNN was implemented based on the deep graph library \citep{wang2019dgl}.  All modules were initialized using random weights. During training, we applied an Adam-Optimizer \citep{kingma2014Adam} starting at an initial learning rate $\epsilon_0=0.0001$. Furthermore, the learning rate was reduced by a factor of five once the evaluation metrics stopped improving for 15 consecutive epochs. The minimum learning rate was set to 
$\epsilon_{min}=10^{-8}$.
The training predictions were supervised on the binary cross entropy loss between estimated and ground  truth edge labels.
Every network architecture was trained for 100 epochs, using multiple GPU nodes simultaneously, each equipped with eight Nvidia Tesla V100s.


\section{Results}\label{Section: Results}

We assess the proposed approaches in comparison to a range of existing methods, using both simulated data and real biological data. In the case of the simulations, we have access to the true, underlying causal graph, and hence can assess results by direct comparison with the ground truth. For the real data examples, we test the model output against the outcome of entirely unseen interventional experiments. 
In all experiments, simulated or real, model output is tested with respect to causal relationships that are entirely unseen in the sense that (i) 
the variable pairs on which the model output is tested are disjoint from those pairs whose causal relationships are provided as inputs during training, and (ii) no data used to define the gold-standard causal relationships against which the model output is tested appear in inputs to the models.

\subsection{Gold-standard simulated benchmark data.}
\begin{figure}[t!]
\centering
\includegraphics[width=0.75\textwidth]{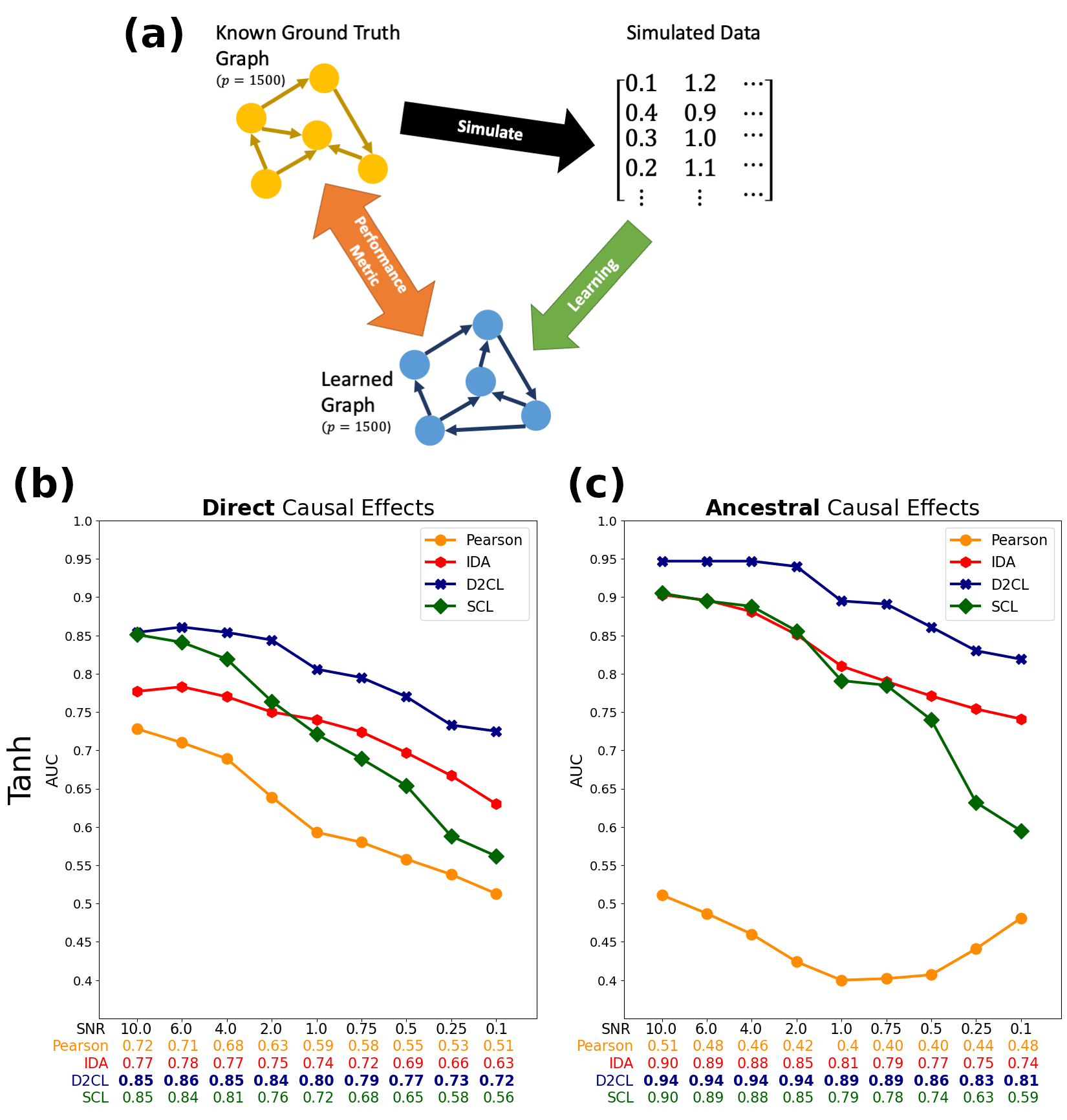}
\caption{Results, simulated data. (a) Overview. Data were simulated from known, gold-standard causal graphs with which the output of the learners was compared. Empirical data were generated using a directed causal graph of specified dimension $p$ using linear and nonlinear structural equation models with noise (see text). 
(b) Results for an illustrative nonlinear case (the tangent hyperbolic), at varying noise levels, for direct causal relationships. Causal area under the ROC-Curve (AUC; with respect 
to the causal ground truth graph) is shown as a function of signal-to-noise ratio (SNR) for an experiment with $p{=}1500$ variables and a sample size of $n{=}1024$. 
D\textsuperscript{2}CL (blue) is compared with: Pearson correlations (yellow; this is a non-causal baseline); IDA (red); and SCL (green). (c) Results for indirect causal relationships, with other settings as in (b). Here, causal AUC is with respect to a graph encoding causal, but potentially indirect, relationships. (Results shown are averages over five data sets at each specified SNR.)}
\label{Figure: Experiment simulations}
\end{figure}

We first tested D\textsuperscript{2}CL using linear and non-linear simulations. These involved generating data  $X$ 
(and obtaining prior knowledge $\Pi$)
from a (linear or non-linear) structural equation model (SEM) with noise, based on a known underlying causal graph $G^*$. 
The  protocol is outlined in Figure \ref{Figure: Experiment simulations}a. In brief, data were generated via structural equations of the form $X_i {=} f_i(Pa_{G^*}(X_i), U_{X_i})$, for $i=1, \ldots, p$, where $p$ is the total number of  variables, $Pa_{G^*}(X_i)$ is the set of parents for node $i$ in the true graph $G^*$, the $U_{X_i}$'s are noise variables (exogenous and jointly independent) and the $f_i$'s functions unknown to the learners. 
Functional forms used include simple linear functions, multi-layer perceptrons (MLPs) with tangent hyperbolic activations, MLPs with leaky ReLU activation, leaky ReLU, a polynomial of order three
and the tangent hyperbolic.
Varying the magnitude of the noise terms allowed us to control the signal-to-noise ratio (SNR), while varying $p$ allowed us to understand the effect of dimensionality. Results were evaluated against the true, gold-standard causal structure $G^*$ and hence tested in causal (and not  correlational or predictive) terms.

Figure \ref{Figure: Experiment simulations}b shows results for a problem of dimension $p{=}1500$ using a nonlinear transition function (the tangent hyperbolic; other functions/configurations are shown in Appendix B) and varying SNR. 
(For these first results, we restricted the dimension of the problem to facilitate comparison 
with approaches that may not scale to larger problems; higher dimensional examples appear below.)
Overall, D\textsuperscript{2}CL 
remains effective across a broad range of SNRs, as well as for a range of linear and nonlinear problems and problem sizes (Appendix B).
These results support the notion that D\textsuperscript{2}CL can learn direct causal edges in systems spanning many variables. 
We note that the comparison with existing  approaches is not one-to-one, since in many cases methods differ in their expected inputs and outputs. For example, IDA is aimed at analysis of  observational data, hence the comparison is unfair since our approach has access also to background causal information $\Pi$. GIES allows for interventional data, but requires different inputs.
Due to these differences in input/output requirements, we emphasize that comparisons here are provided for completeness but with the  caveat that the various methods are intended for different use-cases (and furthermore make assumptions that are likely not met in the real biological data below). 

The  graph $G^*$ in the above examples encodes direct causal relationships since there is an edge from one node to another if the former appears in the equation for the latter. However, in many real-world examples, interest focuses also on indirect effects, that may be mediated by other nodes. For example, if node $A$ has a direct effect on $B$, and $B$ on $C$, intervention on $A$ may change $C$, even though $A$ does not itself appear in the equation for $C$. To study the ability to identify such indirect effects, we next tested the various methods on the task of learning indirect edges. 
This was done in the same way as above, but with the inputs $\Pi$ being indirect edges and output tested against the true indirect graph.

Results appear in Figure \ref{Figure: Experiment simulations}c.  D\textsuperscript{2}CL performs well across a range of SNRs and also in other linear/nonlinear problem configurations 
(Appendix B). 
IDA performs well in case of a linear SEM but not for functions based on nonlinear MLPs. 
These results support the notion that D\textsuperscript{2}CL can learn indirect causal edges over many variables  under conditions of noise and nonlinearity.

\subsection{Large-scale biological data.}
Next, we sought to study performance in the context of real biological data. To this end, we leveraged a large set of gene deletion experiments in yeast \citep{Kemmeren2014}, which have previously been used for causal learning  \citep{Peters2015b,Meinshausen2016,Hill2019}. These data involve measuring gene expression in yeast cells under each of a large number of interventional (gene deletion) experiments. 
 To define causal status, we followed the approach of  \citet{Hill2019}, considering changes under intervention relative to the observational distribution.

In  biological experiments, causal effects may be indirect and our goal in the analysis is to learn a directed graph with nodes corresponding to $p$ observed genes and edges representing (possibly indirect) causal influences. 
Such edges are scientifically interesting as they are relatively amenable to experimental verification  \citep[as noted in][]{Zhang2008CausalGraphs, noe2019}. Cycles can arise in systems biology \citep[see e.g.][]{alon2019} and we do not enforce acyclicity  \citep[see][and references therein, for discussion of cyclic causality]{hyttinen2012}. A fuller discussion of the causal interpretation of laboratory experiments is beyond the scope of this paper, but relevant work includes \citet{eberhardt2007,hyttinen2012,kocaoglu2017} and we direct the interested reader to these references for further discussion.

Since causal background knowledge is an input to our approach, 
it is relevant to consider performance as a function of the amount of such input. To this end, we fixed the problem size to $p=1000$ and varied the number of interventions $m$ whose effects were available to the learner.
Since each experiment involves only a subset of the entire yeast genome,
latent variables 
are present by design. 
The input prior knowledge $\Pi$ is derived from the causal status, but, as in all experiments, is strictly disjoint with respect to any test edges.

\begin{figure}[t!]
\centering
\includegraphics[width=\textwidth]{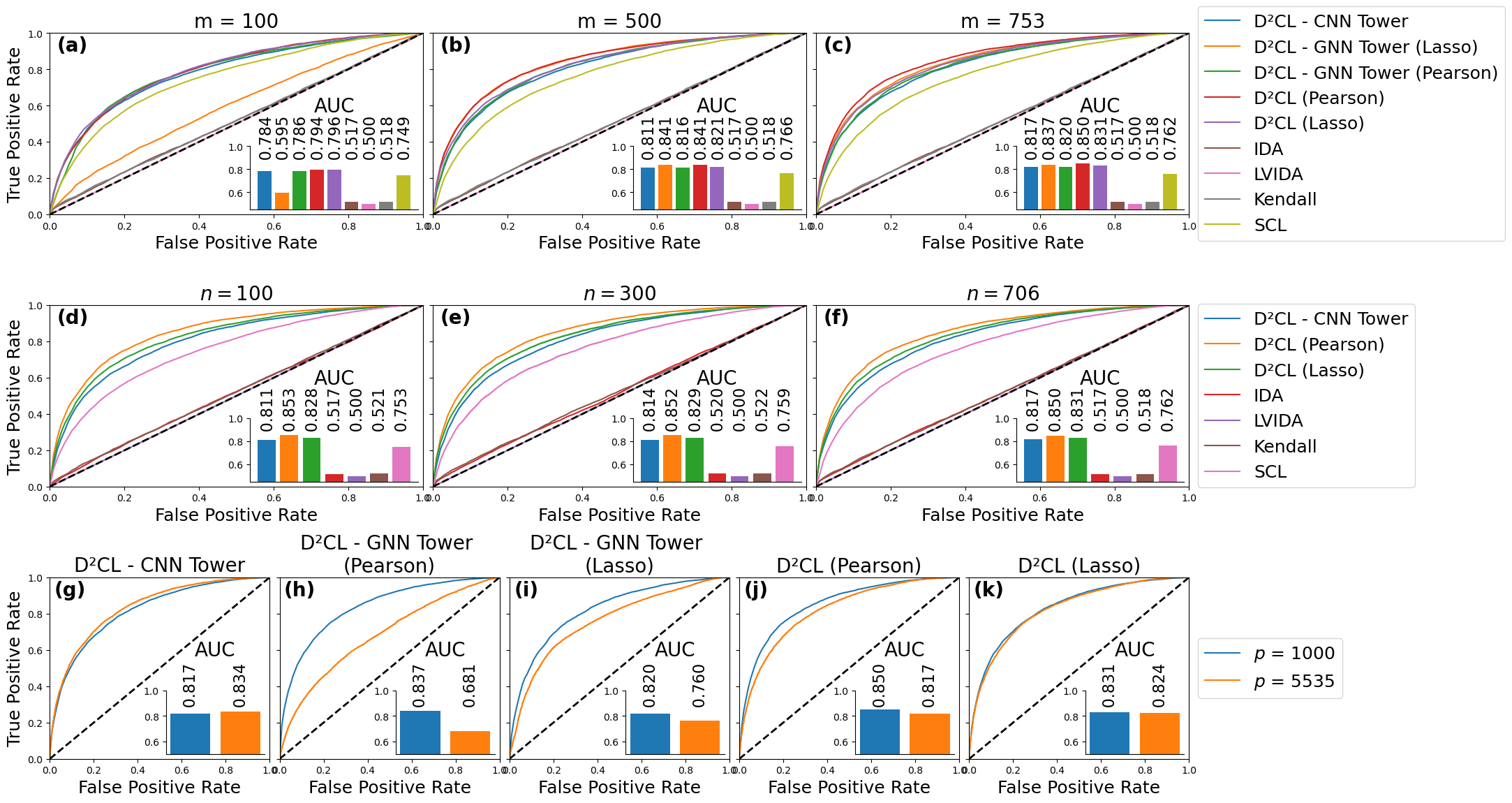}
\caption{Results, biological data. Causal learning methods, including 
D\textsuperscript{2}CL, were applied to gene expression measurements  from yeast cells.
Performance was quantified using causal ROC curves (and the area under the curves, or AUC) computed with respect to a 
causal ground truth obtained from entirely unseen interventional experiments (see text).
Panels (a)--(c):  number of interventions whose effects are available to the learner varied as shown (with problem dimension fixed to $p{=}1000$ and sample size to $n{=}706$).
Panels (d)--(f): 
sample size $n$ varied as shown (with problem dimension fixed to $p{=}1000$ and number of available interventions to $m{=}753$).
Panels (g)--(k): D\textsuperscript{2}CL results for a higher-dimensional setting spanning all available genes with $p{=}5535$ (with $n{=}706$ and $m{=}753$). 
[D\textsuperscript{2}CL variants shown include CNN tower alone, GNN tower alone and the combined architecture;
methods compared against include IDA, LV-IDA, Kendall correlations (as a non-causal baseline) and SCL (see text). For D\textsuperscript{2}CL variants with a GNN component two different initial graph estimates were used based respectively on Pearson correlation coefficients (``Pearson") and on a lightweight  regression (``Lasso"; see text for details).]}
\label{Figure: Experiment yeast}
\end{figure}

Results are shown in Figure \ref{Figure: Experiment yeast}a-c, including the area under the ROC curve (AUC; computed with respect to an experimentally-determined gold-standard, as in \citet{Hill2019}).
Interestingly, the two towers 
differ in some ways: 
the CNN tower degrades slowly with fewer causal inputs while the performance of the GNN tower degrades faster.  GIES \citep{Hauser2012}
 was not effective in this setting (result not shown; findings are in line with \citet{Hill2019} using the same data);
however, we note that GIES requires different inputs to our approach and its assumptions are likely violated in this setting.
Next, to shed light on data efficiency we varied  the sample size $n$ of the data matrix $X$. 
Results are shown in Figure \ref{Figure: Experiment yeast}d-f.

Finally, we tested performance in  a higher dimensional example spanning all $p{=}5535$ available genes (cf.\ Figure \ref{Figure: Experiment yeast}g-k) and found that  D\textsuperscript{2}CL remains effective at   genome scale. Interestingly, while the CNN tower performs particularly well, the GNN tower degrades more. This may be because larger $p$ leads to a larger number of variable pairs (which is helpful for the CNN), but also to a (rapid) increase in the number of nodes and edges in the GNN subgraphs and hence a  harder GNN learning task in practice.

\begin{figure}[t]
\begin{subfigure}[c]{0.5\textwidth}
\centering
\includegraphics[width=\textwidth]{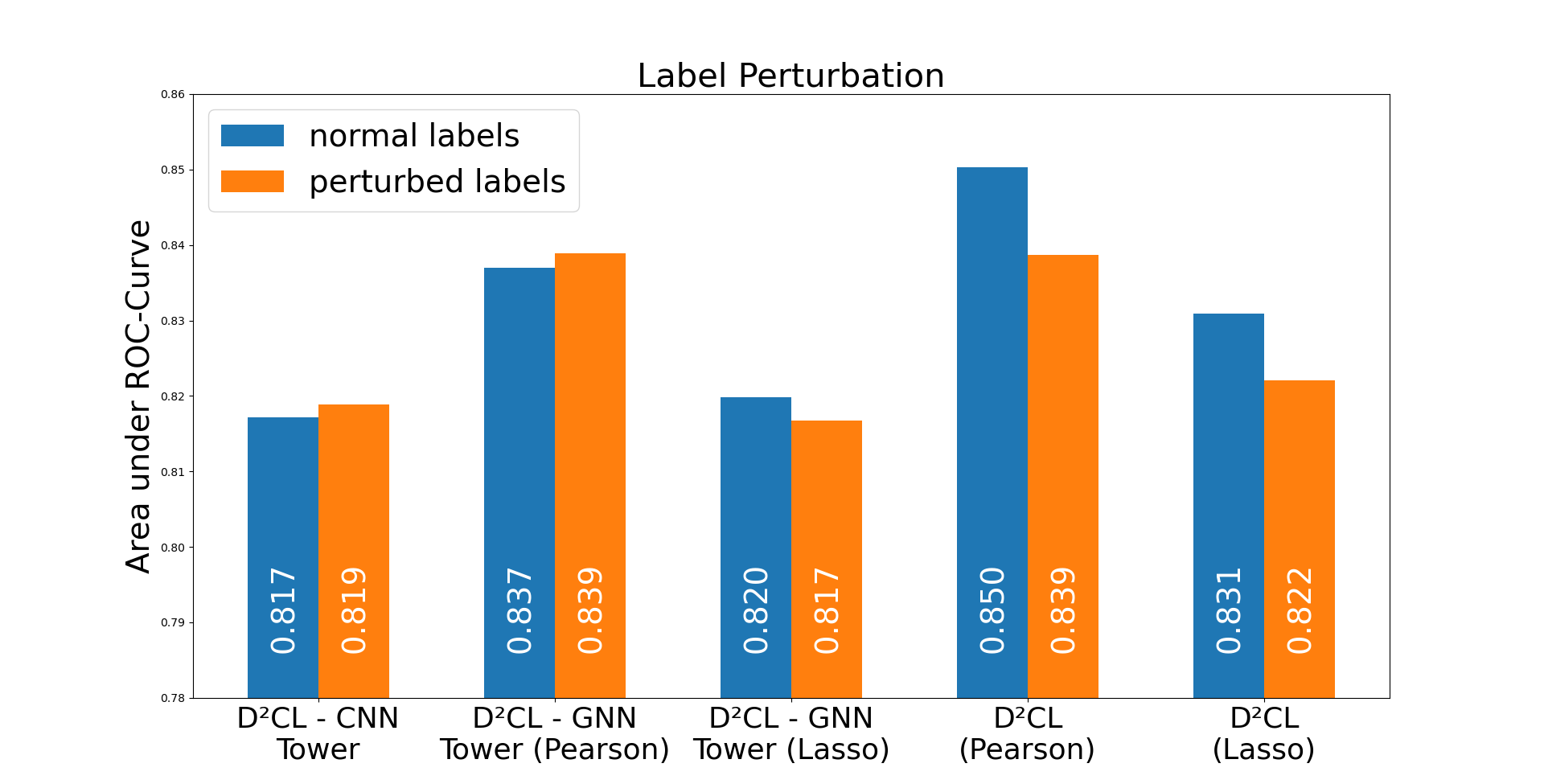}
\subcaption{Label Perturbation} \label{Figure: Label Perturbation}
\end{subfigure}
\begin{subfigure}[c]{0.5\textwidth}
\centering
\includegraphics[width=\textwidth]{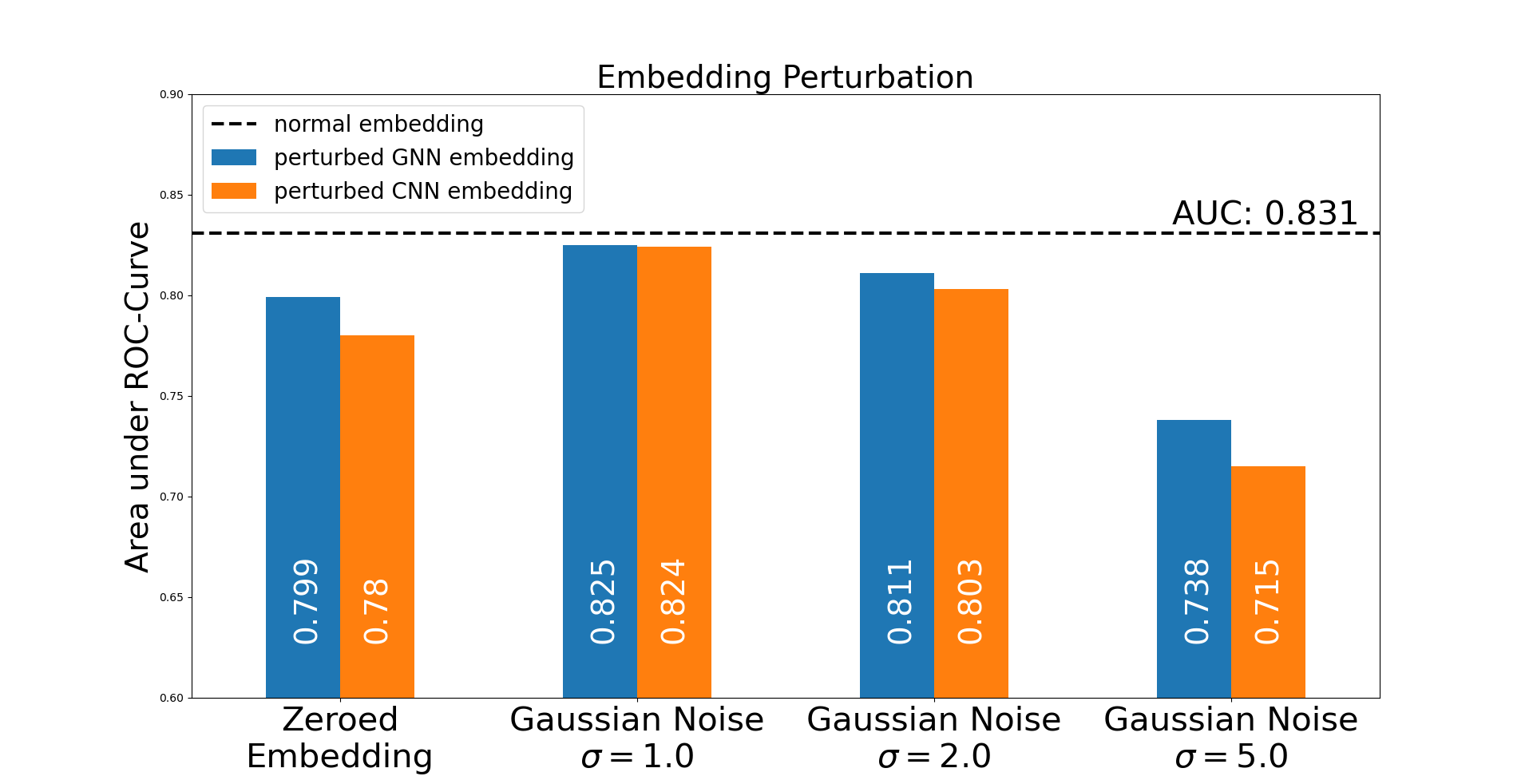}
\subcaption{Embedding Perturbation} \label{Figure: Embedding Perturbation}
\end{subfigure}
\begin{center}
\begin{subfigure}[c]{0.4\textwidth}
\centering
\includegraphics[width=\textwidth]{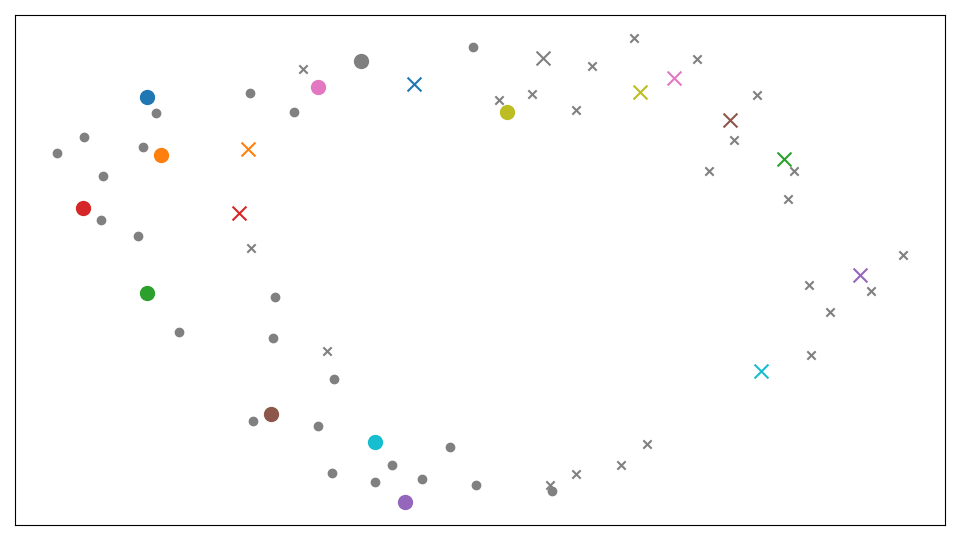}
\subcaption{Low-dimensional representation of feature maps of CNN tower} \label{Figure: Causal Direction ij ji}
\end{subfigure}
\end{center}
\caption{Sensitivity to incorrect causal inputs and additional results on  causal direction.
(a) Robustness to incorrect causal inputs.  Sensitivity of D\textsuperscript{2}CL to errors in  prior/input causal knowledge $\Pi$ was studied by artificially introducing errors into $\Pi$, with 10\% of inputs corrupted (see text). Results quantified via causal AUC (with respect to the correct ground truth).
(b) Ablation-like study in which  failures of either the CNN (orange) or the GNN (blue) tower within D\textsuperscript{2}CL are artificially introduced. The affected embedding is either set to zero or zero-mean Gaussian noise with varying scale is applied. The unaffected case is given as dashed black line. 
(c) Causal direction analysis. 
Low-dimensional representations of latent feature maps of the converged CNN tower at two different layer depths. Edges $A \rightarrow B$ shown as dots and {\it reverse} edges $B \rightarrow A$ as x-shaped markers. An edge and its corresponding reverse is indicated by the same color. For improved readability, ten (randomly chosen) pairs are highlighted in colors and larger markers. 
[D\textsuperscript{2}CL variants include: a CNN tower alone; a GNN tower for two different initial graph estimates; and the complete architecture. Initial graph estimates for the GNN and combined models either 
based on Pearson correlation coefficients (``Pearson") or  a lightweight  regression (``Lasso"; see text).]}
\end{figure}

D\textsuperscript{2}CL leverages prior causal knowledge; however, in practice, available causal inputs $\Pi$ may be {\it incorrect}, e.g.\ due to flawed initial experiments or errors in the known science. To study sensitivity 
to flawed causal inputs 
we introduced errors into $\Pi$. This was done by perturbing 10\% of the inputs (i.e.\ labelling causal pairs as non-causal and vice versa) at the outset.
 Figure \ref{Figure: Label Perturbation} shows 
corresponding  results; 
the networks seem reasonably robust in this sense. 
These experiments point  also to a benefit of the dual network variants: 
when one tower underperforms, the combined network still performs well, as it (automatically) adapts to rely on the effective tower. This aspect is further investigated in Figure \ref{Figure: Embedding Perturbation}.
To test the impact of a failing tower on  overall performance,  the embedding of either tower was modified right before the fusion layer. We considered four different modifications: (i) setting the complete embedding of one tower to zero and hence effectively removing all information from this tower. In the other cases we applied   Gaussian noise with magnitude (ii) $\sigma = 1.0$, (iii) $\sigma=2.0$, and (iv) $\sigma = 5.0$. The results support the notion that even when one tower fails, the second can compensate so that  D\textsuperscript{2}CL still provides useful output.

Causal relations are in general directed and asymmetric.
Given an image representation, the CNN tower extracts feature  maps for (ordered) node pairs. The two-dimensional convolutional operation 
$S(i, j) = \sum_m \sum_n I(m, n)K(i-m, j-n)$
that convolves image $I$ with kernel $K$ would  produce the same feature map for two causal images $I_{k \rightarrow l}$ and $I_{l \rightarrow k}$ if and only if $I_{k \rightarrow l}$ and $I_{l \rightarrow k}$ were identical.
In other words, unless the probability distribution $P(X_i, X_j)$ is perfectly symmetrical around the center of the causal image, the CNN tower can extract causal features that 
differ depending on direction.
Figure \ref{Figure: Causal Direction ij ji} shows a low-dimensional representation of the feature maps of the converged CNN tower; the feature maps differ by direction, supporting the notion that the representations learned are asymmetric.

\section{Conclusions}
Our model leverages deep learning tools to learn causal relationships between variables in a  scalable manner. However, and in contrast to well established approaches based on causal graphical models, it provides only structural output rather than a probability model of the underlying system. 
It would therefore be interesting to consider coupling our approach, as a first  learning step, with a graphical model based analysis in a second step.
This would amount to using the flexible and scalable discriminative approach as a filter to render subsequent causal modelling more tractable.

Despite some initial  ideas presented here (see also Appendix A), 
there remain open questions concerning
the theoretical properties of the kind of approach studied here. In particular, precise conditions on the underlying system needed to ensure that the classification-type approach can guarantee recovery of specific causal  structures remain to be elucidated.
An interesting observation is that the proposed approach may benefit from a ``blessing of dimensionality'', since the learning problem will typically enjoy a larger number of examples as the dimension $p$ grows. Conversely, and in contrast to established statistical-causal models, our approach (at the current stage) {\it cannot} be used in the small-$p$ regime, since then the number of examples will be too small for deep learning.

\bibliographystyle{plainnat}
\bibliography{references}

\begin{thebibliography}{37}
\providecommand{\natexlab}[1]{#1}
\providecommand{\url}[1]{\texttt{#1}}
\expandafter\ifx\csname urlstyle\endcsname\relax
  \providecommand{\doi}[1]{doi: #1}\else
  \providecommand{\doi}{doi: \begingroup \urlstyle{rm}\Url}\fi

\bibitem[Alon(2019)]{alon2019}
Uri Alon.
\newblock \emph{An introduction to systems biology: design principles of
  biological circuits}.
\newblock CRC press, 2019.

\bibitem[Arjovsky et~al.(2019)Arjovsky, Bottou, Gulrajani, and
  Lopez-Paz]{Arjovsky2019}
Martin Arjovsky, Léon Bottou, Ishaan Gulrajani, and David Lopez-Paz.
\newblock Invariant risk minimization.
\newblock \emph{arXiv preprint}, 2019.

\bibitem[Chen et~al.(2019)Chen, Lin, Li, Li, Zhou, and Sun]{chen2019}
Deli Chen, Yankai Lin, Wei Li, Peng Li, Jie Zhou, and Xu~Sun.
\newblock Measuring and relieving the over-smoothing problem for graph neural
  networks from the topological view.
\newblock \emph{Computing Research Repository (CoRR)}, 2019.

\bibitem[Colombo et~al.(2012)Colombo, Maathuis, Kalisch, and
  Richardson]{Colombo2012}
Diego Colombo, Marloes~H. Maathuis, Markus Kalisch, and Thomas~S. Richardson.
\newblock Learning high-dimensional directed acyclic graphs with latent and
  selection variables.
\newblock \emph{The Annals of Statistics}, 40:\penalty0 294--321, 2012.

\bibitem[Eberhardt and Scheines(2007)]{eberhardt2007}
Frederick Eberhardt and Richard Scheines.
\newblock Interventions and causal inference.
\newblock \emph{Philosophy of Science}, 74\penalty0 (5):\penalty0 981--995,
  2007.

\bibitem[Eigenmann et~al.(2020)Eigenmann, Mukherjee, and
  Maathuis]{eigenmann2020}
Marco Eigenmann, Sach Mukherjee, and Marloes Maathuis.
\newblock Evaluation of causal structure learning algorithms via risk
  estimation.
\newblock In \emph{Proceedings of Uncertainty in Artificial Intelligence 2020,
  UAI 2020}, 2020.

\bibitem[Glymour et~al.(2016)Glymour, Pearl, and Jewell]{glymour2016causal}
Madelyn Glymour, Judea Pearl, and Nicholas~P Jewell.
\newblock \emph{Causal inference in statistics: A primer}.
\newblock John Wiley \& Sons, 2016.

\bibitem[Hauser and B{\"u}hlmann(2012)]{Hauser2012}
Alain Hauser and Peter B{\"u}hlmann.
\newblock Characterization and greedy learning of interventional {M}arkov
  equivalence classes of directed acyclic graphs.
\newblock \emph{The Journal of Machine Learning Research}, 13:\penalty0
  2409--2464, 2012.

\bibitem[He et~al.(2015)He, Zhang, Ren, and Sun]{he2015delving}
Kaiming He, Xiangyu Zhang, Shaoqing Ren, and Jian Sun.
\newblock Delving deep into rectifiers: Surpassing human-level performance on
  imagenet classification.
\newblock In \emph{2015 IEEE International Conference on Computer Vision
  (ICCV)}, pages 1026--1034, 2015.

\bibitem[He et~al.(2016)He, Zhang, Ren, and Sun]{he2015deep}
Kaiming He, Xiangyu Zhang, Shaoqing Ren, and Jian Sun.
\newblock Deep residual learning for image recognition.
\newblock In \emph{2016 IEEE Conference on Computer Vision and Pattern
  Recognition (CVPR)}, pages 770--778, 2016.

\bibitem[{Heinze-Deml} et~al.(2018){Heinze-Deml}, {Maathuis}, and
  {Meinshausen}]{Heinze2018}
Christina {Heinze-Deml}, Marloes~H. {Maathuis}, and Nicolai {Meinshausen}.
\newblock {Causal Structure Learning}.
\newblock \emph{Annual Review of Statistics and Its Application}, 5:\penalty0
  371--391, 2018.

\bibitem[Hill et~al.(2016)Hill, Heiser, Cokelaer, et~al.]{Hill2016}
Steven~M. Hill, Laura Heiser, Thomas Cokelaer, et~al.
\newblock Inferring causal molecular networks: Empirical assessment through a
  community-based effort.
\newblock \emph{Nature Methods}, 13:\penalty0 310--318, 2016.

\bibitem[Hill et~al.(2019)Hill, Oates, Blythe, and Mukherjee]{Hill2019}
Steven~M. Hill, Chris~J. Oates, Duncan~A. Blythe, and Sach Mukherjee.
\newblock Causal learning via manifold regularization.
\newblock \emph{The Journal of Machine Learning Research}, 20:\penalty0 1--32,
  2019.

\bibitem[Hyttinen et~al.(2012)Hyttinen, Eberhardt, and Hoyer]{hyttinen2012}
Antti Hyttinen, Frederick Eberhardt, and Patrik~O Hoyer.
\newblock Learning linear cyclic causal models with latent variables.
\newblock \emph{The Journal of Machine Learning Research}, 13\penalty0
  (1):\penalty0 3387--3439, 2012.

\bibitem[Kemmeren et~al.(2014)Kemmeren, Sameith, van~de Pasch, Benschop,
  Lenstra, Margaritis, O~Duibhir, Apweiler, van Wageningen, Ko,
  et~al.]{Kemmeren2014}
Patrick Kemmeren, Katrin Sameith, Loes~AL van~de Pasch, Joris~J Benschop,
  Tineke~L Lenstra, Thanasis Margaritis, Eoghan O~Duibhir, Eva Apweiler, Sake
  van Wageningen, Cheuk~W Ko, et~al.
\newblock Large-scale genetic perturbations reveal regulatory networks and an
  abundance of gene-specific repressors.
\newblock \emph{Cell}, 157\penalty0 (3):\penalty0 740--752, 2014.

\bibitem[Kingma and Ba(2015)]{kingma2014Adam}
Diederik~P. Kingma and Jimmy Ba.
\newblock Adam: A method for stochastic optimization.
\newblock \emph{Computing Research Repository (CoRR)}, abs/1412.6980, 2015.

\bibitem[Kocaoglu et~al.(2017)Kocaoglu, Shanmugam, and
  Bareinboim]{kocaoglu2017}
Murat Kocaoglu, Karthikeyan Shanmugam, and Elias Bareinboim.
\newblock Experimental design for learning causal graphs with latent variables.
\newblock In \emph{Advances in Neural Information Processing Systems 30, NIPS},
  2017.

\bibitem[Li et~al.(2018)Li, Han, and Wu]{Li2018}
Qimai Li, Zhichao Han, and Xiao{-}Ming Wu.
\newblock Deeper insights into graph convolutional networks for semi-supervised
  learning.
\newblock \emph{Computing Research Repository (CoRR)}, abs/1801.07606, 2018.

\bibitem[Lopez-Paz et~al.(2015)Lopez-Paz, Muandet, Sch{\"o}lkopf, and
  Tolstikhin]{lopez2015}
David Lopez-Paz, Krikamol Muandet, Bernhard Sch{\"o}lkopf, and Iliya
  Tolstikhin.
\newblock Towards a learning theory of cause-effect inference.
\newblock In \emph{International Conference on Machine Learning}, 2015.

\bibitem[Maathuis et~al.(2009)Maathuis, Kalisch, and Bühlmann]{Maathuis2009}
Marloes~H. Maathuis, Markus Kalisch, and Peter Bühlmann.
\newblock Estimating high-dimensional intervention effects from observational
  data.
\newblock \emph{The Annals of Statistics}, 37:\penalty0 3133--3164, 2009.

\bibitem[Meinshausen et~al.(2016)Meinshausen, Hauser, Mooij, Peters, Versteeg,
  and Bühlmann]{Meinshausen2016}
Nicolai Meinshausen, Alain Hauser, Joris~M. Mooij, Jonas Peters, Philip
  Versteeg, and Peter Bühlmann.
\newblock Methods for causal inference from gene perturbation experiments and
  validation.
\newblock \emph{Proceedings of the National Academy of Sciences of the United
  States of America}, 113\penalty0 (27):\penalty0 7361--7368, 2016.

\bibitem[Mooij et~al.(2016)Mooij, Peters, Janzing, Zscheischler, and
  Sch{\"{o}}lkopf]{Mooij2016}
Joris~M. Mooij, Jonas Peters, Dominik Janzing, Jakob Zscheischler, and Bernhard
  Sch{\"{o}}lkopf.
\newblock {Distinguishing cause from effect using observational data: Methods
  and benchmarks}.
\newblock \emph{The Journal of Machine Learning Research}, 17:\penalty0 1--102,
  2016.

\bibitem[No{\`e} et~al.(2019)No{\`e}, Taschler, T{\"a}ger, Heutink, and
  Mukherjee]{noe2019}
Umberto No{\`e}, Bernd Taschler, Joachim T{\"a}ger, Peter Heutink, and Sach
  Mukherjee.
\newblock Ancestral causal learning in high dimensions with a human genome-wide
  application.
\newblock \emph{arXiv preprint arXiv:1905.11506}, 2019.

\bibitem[Paszke et~al.(2019)Paszke, Gross, Massa, Lerer, Bradbury, Chanan,
  Killeen, Lin, Gimelshein, Antiga, Desmaison, Kopf, Yang, DeVito, Raison,
  Tejani, Chilamkurthy, Steiner, Fang, Bai, and Chintala]{paszke2019Pytorch}
Adam Paszke, Sam Gross, Francisco Massa, Adam Lerer, James Bradbury, Gregory
  Chanan, Trevor Killeen, Zeming Lin, Natalia Gimelshein, Luca Antiga, Alban
  Desmaison, Andreas Kopf, Edward Yang, Zachary DeVito, Martin Raison, Alykhan
  Tejani, Sasank Chilamkurthy, Benoit Steiner, Lu~Fang, Junjie Bai, and Soumith
  Chintala.
\newblock Pytorch: An imperative style, high-performance deep learning library.
\newblock In \emph{Advances in Neural Information Processing Systems},
  volume~32. Curran Associates, Inc., 2019.

\bibitem[Peters et~al.(2016)Peters, B{\"u}hlmann, and Meinshausen]{Peters2015b}
Jonas Peters, Peter B{\"u}hlmann, and Nicolai Meinshausen.
\newblock Causal inference by using invariant prediction: identification and
  confidence intervals.
\newblock \emph{Journal of the Royal Statistical Society: Series B (Statistical
  Methodology)}, 78\penalty0 (5):\penalty0 947--1012, 2016.

\bibitem[Peters et~al.(2017)Peters, Janzing, and Sch\"olkopf]{Peters2017}
Jonas Peters, Dominik Janzing, and Bernhard Sch\"olkopf.
\newblock \emph{Elements of Causal Inference: Foundations and Learning
  Algorithms}.
\newblock MIT Press, Cambridge, MA, USA, 2017.

\bibitem[Schölkopf et~al.(2012)Schölkopf, Janzing, Peters, Sgouritsa, Zhang,
  and Mooij]{schoelkopf2012causal}
Bernhard Schölkopf, Dominik Janzing, Jonas Peters, Eleni Sgouritsa, Kun Zhang,
  and Joris Mooij.
\newblock On causal and anticausal learning.
\newblock \emph{Proceedings of the 29th International Conference on Machine
  Learning, ICML 2012}, 2, 2012.

\bibitem[Shimizu et~al.(2006)Shimizu, Hoyer, Hyv{\"a}rinen, and
  Kerminen]{Shimizu2006}
Shohei Shimizu, Patrik~O. Hoyer, Aapo Hyv{\"a}rinen, and Antti Kerminen.
\newblock A linear non-{G}aussian acyclic model for causal discovery.
\newblock \emph{The Journal of Machine Learning Research}, 7:\penalty0
  2003--2030, 2006.

\bibitem[Silverman(1986)]{Silverman86}
Bernard~W. Silverman.
\newblock \emph{Density Estimation for Statistics and Data Analysis}.
\newblock Chapman \& Hall, 1986.

\bibitem[Spirtes et~al.(2000)Spirtes, Glymour, and Scheines]{Spirtes2000}
Peter Spirtes, Clark Glymour, and Richard Scheines.
\newblock \emph{Causation, Prediction, and Search}.
\newblock MIT Press, Cambridge, second edition, 2000.
\newblock With additional material by D.\ Heckerman, C.\ Meek, G.F.\ Cooper and
  T.\ Richardson.

\bibitem[Szegedy et~al.(2015)Szegedy, Liu, Jia, Sermanet, Reed, Anguelov,
  Erhan, Vanhoucke, and Rabinovich]{szegedy2014going}
Christian Szegedy, Wei Liu, Yangqing Jia, Pierre Sermanet, Scott Reed, Dragomir
  Anguelov, Dumitru Erhan, Vincent Vanhoucke, and Andrew Rabinovich.
\newblock Going deeper with convolutions.
\newblock In \emph{2015 IEEE Conference on Computer Vision and Pattern
  Recognition (CVPR)}, 2015.

\bibitem[Turlach(1993)]{turlach1993bandwidth}
Berwin~A Turlach.
\newblock Bandwidth selection in kernel density estimation: A review.
\newblock In \emph{CORE and Institut de Statistique}, 1993.

\bibitem[Wang et~al.(2019)Wang, Zheng, Ye, Gan, Li, Song, Zhou, Ma, Yu, Gai,
  Xiao, He, Karypis, Li, and Zhang]{wang2019dgl}
Minjie Wang, Da~Zheng, Zihao Ye, Quan Gan, Mufei Li, Xiang Song, Jinjing Zhou,
  Chao Ma, Lingfan Yu, Yu~Gai, Tianjun Xiao, Tong He, George Karypis, Jinyang
  Li, and Zheng Zhang.
\newblock Deep graph library: A graph-centric, highly-performant package for
  graph neural networks.
\newblock \emph{arXiv preprint arXiv:1909.01315}, 2019.

\bibitem[Xie et~al.(2017)Xie, Girshick, Dollár, Tu, and He]{xie2017aggregated}
Saining Xie, Ross Girshick, Piotr Dollár, Zhuowen Tu, and Kaiming He.
\newblock Aggregated residual transformations for deep neural networks.
\newblock In \emph{2017 IEEE Conference on Computer Vision and Pattern
  Recognition (CVPR)}, 2017.

\bibitem[Zhang(2008)]{Zhang2008CausalGraphs}
Jiji Zhang.
\newblock {Causal reasoning with ancestral graphs}.
\newblock \emph{The Journal of Machine Learning Research}, 9:\penalty0
  1437--1474, 2008.

\bibitem[Zhang and Chen(2018)]{zhang2018}
Muhan Zhang and Yixin Chen.
\newblock Link prediction based on graph neural networks.
\newblock In \emph{Advances in Neural Information Processing Systems 2018,
  NeurIPS 2018}, 2018.

\bibitem[Zhang et~al.(2018)Zhang, Cui, Neumann, and Chen]{Zhang2018AnED}
Muhan Zhang, Zhicheng Cui, M.~Neumann, and Yixin Chen.
\newblock An end-to-end deep learning architecture for graph classification.
\newblock In \emph{AAAI}, 2018.

\end{thebibliography}

\newpage
\appendix
\section*{Appendix A: Causal interpretation of the learning scheme}
\label{appendix:interpretation}

Here, we provide some intuition on why discriminative learning can be effective in the setting of interest here. We note that the following arguments are not intended to constitute a  theory at this stage but rather to help gain understanding of the 
conditions under which discriminative causal structure learning as described in the Main Text may be expected to be effective. 


We start with a general causal framework and then introduce 
assumptions for D\textsuperscript{2}CL ({\it MGA} and {\it DCSI}, see below).
Following \citet{Peters2017, schoelkopf2012causal}, we assume decomposition of the underlying system into modular and independent mechanisms:

{\it Independent Causal Mechanisms (ICMs):} The causal generative process of a system's variables is composed of autonomous modules that do not inform or influence each other.

For   variables $X_i$ assume a structural causal model with equations
$X_i{=}f_i(Pa_{G^*}(X_i), U_{X_i}), i=1,\ldots,p$, 
where $Pa_{G^*}(X_i)$ denotes the set of parents in the ground truth graph $G^*$ for node $i$ and $f_i$ is a node-specific function. Exogenous noise terms $U_{X_i}$ are assumed jointly independent and distributed as $U_{X_i} {\sim} p_i$, where $p_i$ is a node-specific density.

Our approach treats the $f_i$'s and $p_i$'s as unknown but assumes they are related at a higher level. This can be formalized as a {\it meta-generator assumption} 
as follows:

{\it Meta-Generator Assumption (MGA):} 
For a specific system $W$, the functions $f_i$ and noise distributions $p_i$ are (independently) generated as 
 $f_i {\sim} \mathcal{F}_W$ and $p_i {\sim} \mathcal{P}_W$, where $\mathcal{F}_W$ denotes a {\it function generator}, and $\mathcal{P}_W$ a {\it stochastic generator}, that are specific to the applied problem setting $W$.

MGA is motivated by the notion that in any particular  real-world  system, underlying (biological, physical, social, etc.) processes tend to share some functional and stochastic aspects, which impart some higher-level regularity. That is, MGA states that in a given applied context, functions $f_i$ and noise terms $U_{X_i}$ while unknown, varied and potentially complex, are nonetheless related at a ``meta''-level.  The generators $\mathcal{F}_W,\mathcal{P}_W$ are random processes, representing respectively a ``distribution over functions" and 
``distribution over distributions", whose role here is to capture the notion of relatedness among $f_i$'s (respectively $p_i$'s) in a given setting $W$.
Note that $\mathcal{F}_W,\mathcal{P}_W$ are treated as unknown and never directly estimated (see below).

As noted above, we focus on the causal status of variable pairs $(X_i,X_j)$ (rather than general tuples) which is the simplest possible case under MGA. 
Furthermore, in both our work and the majority of interventional studies in applications such as biology, single interventions (rather than joint interventions on multiple nodes) are the norm.
Focusing on single interventions motivates the following additional assumption:

{\it Dominant cause under single interventions (DCSI):} A sufficiently large change in one of potentially multiple causes leads to a change w.r.t.\ the effect. Therefore, single interventions are sufficient to drive variation in the child distribution.

{\it From MGA and DCSI to discriminative causal structure learning}. 
Consider an applied problem $W$ with underlying causal graph $G^*_W$, treated as fixed but unknown. The associated functions and noise terms are also unknown but assumed to follow MGA. Then, under DCSI, we have  that {\it all} pairs of the form $(X_i,X_j)$ , have underlying relationships of the form $X_j {=} f_j(X_i, U_{X_j})$  with components following the MGA (i.e.\ drawn from generators 
$\mathcal{F}_W,\mathcal{P}_W$). 
This in turn suggests that within the setting $W$, identification of causal pairs can be treated as a classification problem, since all pairs share the {\it same} generators. In other words, MGA restricts the distribution over relations of variables and noise terms to system-specific distributions.

Note that no particular assumption is made on the individual functions $f_i$, only that they are mutually related on a higher level. Furthermore, the generators themselves need not to be known or are directly estimated, it is only important that they are shared across the applied setting $W$.
Note that a model learned for setting $W$ will not in general be able to classify pairs in an entirely different applied setting $W'$ (since the generators may then differ strongly), i.e.\ we do not seek to learn ``universal" patterns that apply to {\it all} causal relations in any system whatsoever.  The classification task of D\textsuperscript{2}CL aims at telling apart causal relationships, related by the system-specific function generator $F_W$, from non-causal ones. 
We note that in real systems, $f_i$'s may be coupled via constraints on global functionality, hence non-independent, 
however, the good performance seen in the Main Text empirically justifies the approach. We emphasize that while the ideas above provide some initial intuition, further work is needed to better understand the  properties of the kind of approach studied here from a theoretical point of view.

\section*{Appendix B: additional results}
\label{appendix:results}

{\bf Simulated data: direct causal relationships}
\begin{table*}[htbp]
\centering
\caption{AUC values for direct cause-effect relations for $p=|V|=1500$.}
\label{Table: DAG}
\begin{adjustbox}{max width=\textwidth}
\begin{tabular}{l|l|l|l|l|l|l|l|l|l|l|l|l|l|l|l|l|l|l|l|l|l|l|l|l} 
\toprule
\multicolumn{1}{c|}{} & \multicolumn{4}{c|}{Linear}              & \multicolumn{4}{c|}{MLP(tanh)}                  & \multicolumn{4}{c|}{MLP(leaky ReLU)}     & \multicolumn{4}{c|}{Tanh}                & \multicolumn{4}{c|}{Leaky ReLU}          & \multicolumn{4}{c}{Polynom 3}                      \\
SNR                   & Pearson & IDA   & D\textsuperscript{2}CL           & SCL   & Pearson        & IDA   & D\textsuperscript{2}CL           & SCL   & Pearson & IDA   & D\textsuperscript{2}CL           & SCL   & Pearson & IDA   & D\textsuperscript{2}CL           & SCL   & Pearson & IDA   & D\textsuperscript{2}CL           & SCL   & Pearson & IDA            & D\textsuperscript{2}CL           & SCL    \\ 
\hline
10.00                 & 0.718   & 0.748 & \textbf{0.789} & 0.641 & \textbf{0.691} & 0.625 & 0.686          & 0.634 & 0.688   & 0.608 & \textbf{0.693} & 0.623 & 0.728   & 0.777 & \textbf{0.854} & 0.851 & 0.756   & 0.837 & \textbf{0.843} & 0.729 & 0.809   & \textbf{0.848} & 0.829          & 0.641  \\
6.00                  & 0.700   & 0.771 & \textbf{0.795} & 0.617 & \textbf{0.670} & 0.609 & 0.658          & 0.638 & 0.666   & 0.590 & \textbf{0.683} & 0.602 & 0.710   & 0.783 & \textbf{0.861} & 0.841 & 0.736   & 0.818 & \textbf{0.839} & 0.692 & 0.784   & \textbf{0.824} & 0.821          & 0.637  \\
4.00                  & 0.684   & 0.768 & \textbf{0.784} & 0.616 & \textbf{0.648} & 0.590 & 0.647 & 0.625 & 0.652   & 0.562 & \textbf{0.667} & 0.584 & 0.689   & 0.770 & \textbf{0.854} & 0.819 & 0.700   & 0.792 & \textbf{0.831} & 0.668 & 0.735   & 0.787          & \textbf{0.812} & 0.628  \\
2.00                  & 0.638   & 0.781 & \textbf{0.802} & 0.615 & 0.613          & 0.544 & \textbf{0.639} & 0.594 & 0.617   & 0.521 & \textbf{0.661} & 0.592 & 0.639   & 0.750 & \textbf{0.844} & 0.764 & 0.644   & 0.743 & \textbf{0.815} & 0.630 & 0.651   & 0.722          & \textbf{0.777} & 0.617  \\
1.00                  & 0.595   & 0.774 & \textbf{0.796} & 0.614 & 0.572          & 0.506 & \textbf{0.622} & 0.551 & 0.575   & 0.487 & \textbf{0.642} & 0.546 & 0.593   & 0.740 & \textbf{0.806} & 0.721 & 0.582   & 0.701 & \textbf{0.787} & 0.619 & 0.552   & 0.659          & \textbf{0.743} & 0.598  \\
0.75                  & 0.589   & 0.765 & \textbf{0.793} & 0.612 & 0.556          & 0.494 & \textbf{0.619} & 0.566 & 0.567   & 0.483 & \textbf{0.638} & 0.568 & 0.580   & 0.724 & \textbf{0.795} & 0.689 & 0.572   & 0.696 & \textbf{0.783} & 0.610 & 0.539   & 0.645          & \textbf{0.734} & 0.603  \\
0.50                  & 0.558   & 0.748 & \textbf{0.787} & 0.610 & 0.536          & 0.473 & \textbf{0.631} & 0.540 & 0.544   & 0.459 & \textbf{0.641} & 0.567 & 0.558   & 0.697 & \textbf{0.770} & 0.654 & 0.548   & 0.678 & \textbf{0.771} & 0.606 & 0.521   & 0.640          & \textbf{0.717} & 0.592  \\
0.25                  & 0.537   & 0.735 & \textbf{0.784} & 0.572 & 0.530          & 0.467 & \textbf{0.617} & 0.517 & 0.496   & 0.434 & \textbf{0.624} & 0.552 & 0.538   & 0.667 & \textbf{0.733} & 0.588 & 0.517   & 0.667 & \textbf{0.748} & 0.579 & 0.514   & 0.634          & \textbf{0.694} & 0.543  \\
0.10                  & 0.523   & 0.730 & \textbf{0.774} & 0.558 & 0.492          & 0.441 & \textbf{0.618} & 0.530 & 0.507   & 0.439 & \textbf{0.616} & 0.528 & 0.513   & 0.630 & \textbf{0.725} & 0.562 & 0.503   & 0.661 & \textbf{0.743} & 0.559 & 0.492   & 0.620          & \textbf{0.691} & 0.539  \\
\bottomrule
\end{tabular}
\end{adjustbox}
\end{table*}

{\bf Simulated data: indirect/ancestral relationships}
\begin{table*}[htbp]
\centering
\caption{AUC values for indirect cause-effect relations for $p=|V|=1500$.}
\label{Table: AncestralDAG}
\begin{adjustbox}{max width=\textwidth}
\begin{tabular}{l|l|l|l|l|l|l|l|l|l|l|l|l|l|l|l|l|l|l|l|l|l|l|l|l} 
\toprule
      & \multicolumn{4}{c|}{Linear}              & \multicolumn{4}{c|}{MLP(tanh)}           & \multicolumn{4}{c|}{MLP(leaky ReLU)}     & \multicolumn{4}{c|}{Tanh}                & \multicolumn{4}{c|}{Leaky ReLU}          & \multicolumn{4}{c}{Polynom 3}             \\
SNR   & Pearson & IDA   & D\textsuperscript{2}CL           & SCL   & Pearson & IDA   & D\textsuperscript{2}CL           & SCL   & Pearson & IDA   & D\textsuperscript{2}CL           & SCL   & Pearson & IDA   & D\textsuperscript{2}CL           & SCL   & Pearson & IDA   & D\textsuperscript{2}CL           & SCL   & Pearson & IDA   & D\textsuperscript{2}CL           & SCL    \\ 
\hline
10.00 & 0.553   & 0.907 & \textbf{0.928} & 0.708 & 0.548   & 0.522 & \textbf{0.733} & 0.700 & 0.563   & 0.483 & \textbf{0.789} & 0.738 & 0.511   & 0.903 & \textbf{0.947} & 0.905 & 0.502   & 0.857 & \textbf{0.943} & 0.839 & 0.610   & 0.822 & \textbf{0.933} & 0.761  \\
6.00  & 0.540   & 0.905 & \textbf{0.925} & 0.700 & 0.537   & 0.502 & \textbf{0.720} & 0.658 & 0.552   & 0.458 & \textbf{0.775} & 0.735 & 0.487   & 0.896 & \textbf{0.947} & 0.895 & 0.501   & 0.852 & \textbf{0.941} & 0.808 & 0.598   & 0.815 & \textbf{0.927} & 0.751  \\
4.00  & 0.530   & 0.905 & \textbf{0.928} & 0.677 & 0.533   & 0.490 & \textbf{0.711} & 0.675 & 0.548   & 0.447 & \textbf{0.767} & 0.727 & 0.460   & 0.881 & \textbf{0.947} & 0.888 & 0.504   & 0.848 & \textbf{0.937} & 0.782 & 0.581   & 0.803 & \textbf{0.914} & 0.732  \\
2.00  & 0.506   & 0.897 & \textbf{0.928} & 0.619 & 0.523   & 0.461 & \textbf{0.683} & 0.656 & 0.532   & 0.420 & \textbf{0.766} & 0.695 & 0.424   & 0.851 & \textbf{0.940} & 0.856 & 0.501   & 0.834 & \textbf{0.920} & 0.736 & 0.543   & 0.775 & \textbf{0.879} & 0.704  \\
1.00  & 0.507   & 0.888 & \textbf{0.925} & 0.609 & 0.513   & 0.433 & \textbf{0.660} & 0.626 & 0.518   & 0.393 & \textbf{0.738} & 0.672 & 0.400   & 0.810 & \textbf{0.895} & 0.791 & 0.502   & 0.824 & \textbf{0.900} & 0.676 & 0.520   & 0.753 & \textbf{0.831} & 0.658  \\
0.75  & 0.507   & 0.882 & \textbf{0.920} & 0.631 & 0.513   & 0.431 & \textbf{0.638} & 0.610 & 0.514   & 0.385 & \textbf{0.721} & 0.637 & 0.402   & 0.790 & \textbf{0.891} & 0.785 & 0.501   & 0.822 & \textbf{0.888} & 0.667 & 0.516   & 0.747 & \textbf{0.820} & 0.641  \\
0.50  & 0.506   & 0.877 & \textbf{0.918} & 0.505 & 0.510   & 0.422 & \textbf{0.618} & 0.577 & 0.513   & 0.387 & \textbf{0.713} & 0.655 & 0.407   & 0.771 & \textbf{0.861} & 0.740 & 0.502   & 0.821 & \textbf{0.880} & 0.651 & 0.510   & 0.742 & \textbf{0.808} & 0.635  \\
0.25  & 0.513   & 0.873 & \textbf{0.913} & 0.516 & 0.511   & 0.423 & \textbf{0.622} & 0.567 & 0.509   & 0.385 & \textbf{0.703} & 0.639 & 0.441   & 0.754 & \textbf{0.830} & 0.632 & 0.498   & 0.816 & \textbf{0.861} & 0.624 & 0.502   & 0.736 & \textbf{0.792} & 0.589  \\
0.10  & 0.506   & 0.867 & \textbf{0.905} & 0.542 & 0.501   & 0.417 & \textbf{0.617} & 0.571 & 0.503   & 0.385 & \textbf{0.705} & 0.613 & 0.481   & 0.741 & \textbf{0.819} & 0.595 & 0.500   & 0.817 & \textbf{0.865} & 0.598 & 0.503   & 0.734 & \textbf{0.784} & 0.537  \\
\bottomrule
\end{tabular}
\end{adjustbox}
\end{table*}

\end{document}